%% file: main.tex
\pdfoutput=1

\documentclass[11pt,dvipsnames]{article}

\usepackage{emnlp2022}

\usepackage{times}
\usepackage{latexsym}

\usepackage[T1]{fontenc}

\usepackage[utf8]{inputenc}

\usepackage{microtype}

\usepackage{mathtools}
\usepackage{graphicx}
\usepackage{cleveref}
\usepackage{booktabs}
\usepackage{caption}
\usepackage{subcaption}
\usepackage{placeins}
\usepackage{textcomp, xspace}
\usepackage{chngcntr}
\usepackage{nth}

\usepackage{arydshln}
\usepackage{multirow}
\usepackage{amssymb}
\usepackage{pifont}
\newcommand{\cmark}{\ding{51}}
\newcommand{\xmark}{\ding{55}}

\newcommand{\bignews}{\textsc{BigNewsBln}\xspace}

\newcommand{\bnaimg}{\textsc{BNA-Img}\xspace}
\newcommand{\bnimgcap}{\textsc{BN-ImgCap}\xspace}

\newcommand\blfootnote[1]{
    \begingroup
    \renewcommand\thefootnote{}\footnote{#1}
    \addtocounter{footnote}{-1}
    \endgroup
}

\title{Late Fusion with Triplet Margin Objective for \\ Multimodal Ideology Prediction and Analysis}

\author{Changyuan Qiu* \quad Winston Wu* \quad Xinliang Frederick Zhang \quad Lu Wang \\
  Computer Science and Engineering \\
  University of Michigan \\
  \texttt{\{peterqiu,wuws,xlfzhang,wangluxy\}@umich.edu}
}

\begin{document}
\maketitle

\begin{abstract}

Prior work on ideology prediction has largely focused on single modalities, i.e., text or images.
In this work, we introduce the task of \textit{multimodal ideology prediction}, where a model predicts binary or five-point scale ideological leanings, given a text-image pair with political content.
We first collect five new large-scale datasets with English documents and images along with their ideological leanings, covering news articles from a wide range of US mainstream media and social media posts from Reddit and Twitter.
We conduct in-depth analyses of news articles and reveal differences in image content and usage across the political spectrum.
Furthermore, we perform extensive experiments and ablation studies, demonstrating the effectiveness of targeted pretraining objectives on different model components. Our best-performing model, a late-fusion architecture pretrained with a triplet objective over multimodal content, outperforms the state-of-the-art text-only model by almost 4\% and a strong multimodal baseline with no pretraining by over 3\%.

\end{abstract}

\blfootnote{* Equal contribution.}

\input{sections/01-introduction}

\input{sections/02-related}

\input{sections/03-data}

\input{sections/04-models}

\input{sections/05-experiments}

\input{sections/06-results}

\input{sections/07-conclusion}

\section*{Acknowledgments}

This work is supported in part by the National Science Foundation under grant III-2127747 and by the Air Force Office of Scientific Research through grant FA9550-22-1-0099.
We would like to thank the members of the LAUNCH lab at the University of Michigan and the anonymous reviewers for their helpful comments and feedback. We would especially like to thank Yujian Liu for providing us with the BigNews dataset.

\input{sections/08-ethics}

\input{sections/09-limitations}

\FloatBarrier
\bibliography{anthology,custom}
\bibliographystyle{acl_natbib}

\appendix
\counterwithin{figure}{section}
\counterwithin{table}{section}

\input{sections/A-appendix}

\end{document}

%% file: sections/01-introduction.tex
\section{Introduction}

In an increasingly divided world rife with misinformation and hyperpartisanship, it is important to understand the perspectives and biases of the creators of the media that we consume. Media bias can manifest in many ways and has been analyzed from a variety of angles:
the news may favor one side of a political issue \citep{card-etal-2015-media,mendelsohn-etal-2021-modeling},
select certain events to report on \citep{mccarthy1996images,oliver2000political,fan-etal-2019-plain} or even distort or misrepresent facts \citep{gentzkow2006media,entman2007framing}.

Identifying a news article's underlying political slant is the task of \textit{ideology prediction}, which has focused largely on political texts like news articles and has been tackled with a variety of models, including Bayesian approaches with attention \citep{kulkarni-etal-2018-multi}, graph neural networks \citep{li-goldwasser-2019-encoding}, and LSTMs and BERT \citep{baly-etal-2020-detect}. However, past work focuses solely on news articles' \textit{text}; news articles contain other forms of non-verbal information in which the underlying ideology may be realized.

\begin{figure}
    \centering
    \begin{subfigure}[b]{0.40\columnwidth}
    \includegraphics[width=\columnwidth]{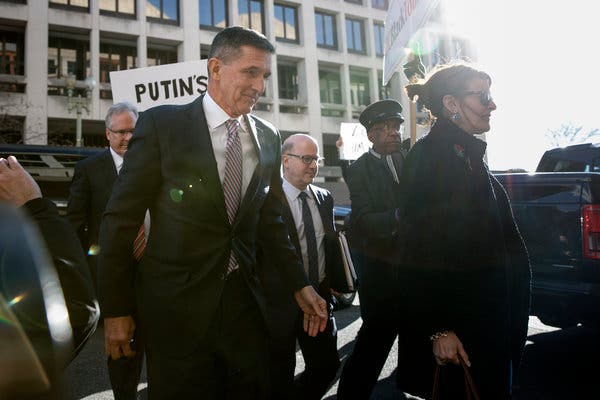}
    \caption{}
    \end{subfigure}
    ~
    \begin{subfigure}[b]{0.44\columnwidth}
    \includegraphics[width=\columnwidth]{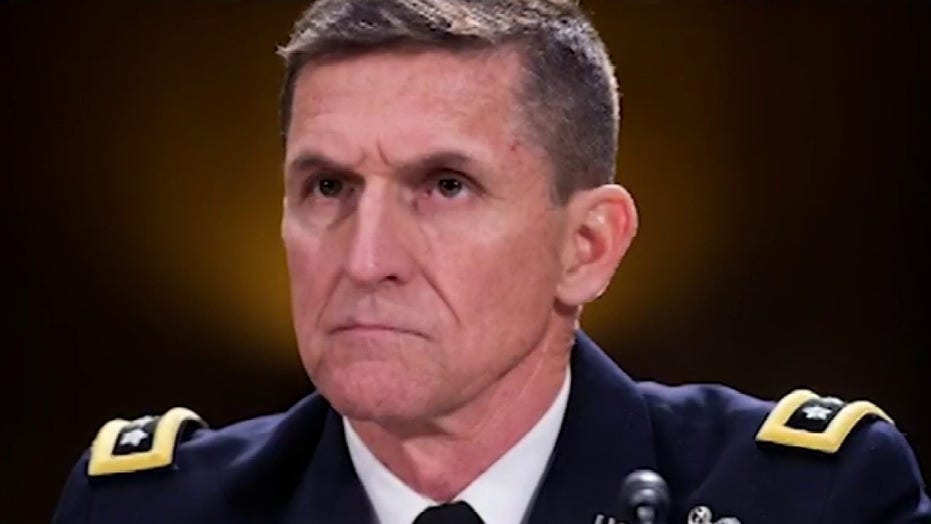}
    \caption{}
    \end{subfigure}

    \caption{Two images from separate sources depicting \textit{Federal Judge Pauses Justice Department Effort to Dismiss Michael Flynn Case}. In (a), from New York Times, Flynn is shown with several other figures and has a positive expression. In contrast, in (b), from Fox News, Flynn is the sole figure, with a negative expression.}

    \label{fig:flynn-comparison}
    \vspace{-1mm}
\end{figure}

Consider \Cref{fig:flynn-comparison}a and \Cref{fig:flynn-comparison}b, two images from articles depicting the same news story, but by news sources with opposing ideologies (New York Times and Fox News, respectively). The underlying ideology of the news source may influence the choice of image: in \Cref{fig:flynn-comparison}a, Michael Flynn is depicted with a happy expression and surrounded by other figures, while in \Cref{fig:flynn-comparison}b, Flynn bears a stern expression and is the sole figure. Images are an integral part of news articles: over 56\% of articles in AllSides\footnote{\url{allsides.com}, a website that categorizes media outlets and news articles by political slant. It associates multiple \textit{articles} with a single \textit{story} about which these articles were written.} include at least one image. Images are often used to frame certain issues or influence the reader's opinion. For example, liberal websites devote more visual coverage of Donald Trump and also portray Trump with more negative emotions compared with conservative websites \citep{boxell2021slanted}.
In addition, images of groups of immigrants, in contrast to individual immigrants, tend to decrease survey respondents' support for immigration \citep{madrigal2021migrants}.
These findings naturally lead us to conduct a study of political images. In \Cref{sec:image-analysis}, we present a thorough analysis of images, finding, \textit{inter alia}, that (1) liberal sources tend to include more figures in an image, (2) conservative sources have a higher usage of collage images, and (3) faces are more likely to show negative or neutral emotion rather than positive.

Although modern American politics have centered around two polar opposites \citep{klein2020we}, 38\% of US adults identify as politically independent and do not agree wholly with left or right ideologies.\footnote{\tiny \url{https://www.pewresearch.org/fact-tank/2019/05/15/facts-about-us-political-independents/}}  Ideology exists on a spectrum \citep{bobbio1996left}, and we wish to predict more fine-grained ideology than merely left or right.
Thus, we define \textit{multimodal ideology prediction} in this work as predicting one of five ideological slants (left, lean left, center, lean right, right) given both an article's text and cover image. To support this new task, we present several new large-coverage datasets of news articles and images across the ideological spectrum from various sources including AllSides, Reddit, Twitter, and 11 independent news sources.

We experiment with several early and late fusion architectures and
evaluate several continued pretraining objectives to improve the image and text encoder separately as well as jointly.
Our technical contributions include a novel triplet margin loss over multimodal inputs, and the first systematic study on multimodal models for ideology prediction, which reveals several findings:
(1) images are indeed helpful for ideology prediction, improving over a text-only model especially on right-leaning images;
(2) late-fusion architectures perform better than early-fusion architectures; (3) ideology-driven pretraining on both the text and image encoders is beneficial; (4) fine-tuning with a \textit{joint triplet difference loss} encourages the model to learn informative representations of ideology.
Code and datasets can be found at \url{github.com/launchnlp/mulmodide}.

%% file: sections/02-related.tex
\section{Related Work}

\paragraph{Media Bias/Ideology on Texts} The study of media bias and ideology has a long history going as far back as \citet{white1950gate}. Computationally, researchers have studied various approaches in classical machine learning as well as neural methods \citep[e.g.][]{evans2007recounting,yu2008classifying,sapiro2019examining,iyyer-etal-2014-political}. However, these works focus solely on text. There exist several resources of news articles across the political spectrum, compiled for the purpose of educating users on media bias \citep[][\footnote{\url{https://adfontesmedia.com}},\footnote{\url{https://www.allsides.com}},\footnote{\url{https://mediabiasfactcheck.com}}]{park2009newscube,park2011newscube,hamborg2017matrix}. Multimodal studies such as ours need annotated data for training and testing. Thus, we collect several datasets containing both political text and images from various sources.

\paragraph{Media Bias/Ideology on Texts and Images} Only very recently has there been much study on media bias with respect to both text and images. Existing work on characterizing political images has been limited to narrow domains such as Russian trolls on Twitter \citep{zannettou2020characterizing}, political memes \citep{beskow2020evolution}, and COVID content on TikTok \citep{southwick2021characterizing}. In addition, data containing both text and images annotated for political ideology are not readily available.\footnote{\citet{NEURIPS2019_e4dd5528} claims to have released such a dataset, but their link was dead. Their dataset also does not annotate fine-grained ideology.} Thus, we collect, annotate, and analyze a variety of new datasets, focusing on political figures in news images. For the tasks of multimodal ideology prediction, one similar work to ours is \citet{NEURIPS2019_e4dd5528}, who investigate adding text to help an image encoder train an enhanced representation of images. Afterwards, they ignore the text and focus on ideology prediction from images alone. They consider only left or right ideologies, in contrast to our more fine-grained 5-way set.

%% file: sections/03-data.tex
\section{Data}

In this section, we describe several datasets collected in this work for pretraining and finetuning the proposed models.

\subsection{Pretraining Datasets}

We build two pretraining datasets based on \bignews \citep{liu2022politics}, a corpus of over 1.9M English news articles collected from 11 news sources balanced across the political spectrum.

\paragraph{\bnimgcap} We first collect a new dataset, \bnimgcap, of 1.2M images that occur anywhere in a news article,\footnote{In contrast to just the cover image.} along with their captions, from seven news sources represented in \bignews, chosen to roughly cover equal proportions of left-, center-, and right-leaning ideologies.
Details of this collection process are described in \Cref{sec:image-collection}. We use these image-caption pairs in our experiments for pretraining the image encoder with the InfoNCE loss and bidirectional captioning loss (\Cref{sec:pretraining}).

\paragraph{\bnaimg} \citet{liu2022politics} also introduced a subset of \bignews called \textsc{BigNewsAlign} containing articles associated with a story cluster, i.e., news articles from different news sources but written about the same story, for pretraining with an ideological triplet loss. From this subset, we identify articles containing images and crawl these images from each article's corresponding webpage. We call this dataset of article text and images \bnaimg and use this for pretraining the cross-modal attention with our proposed triplet margin loss (\Cref{sec:pretraining}). \Cref{tab:img-data-summary} summarizes these datasets.

\begin{table*}[t]
    \centering
    \tiny

    \begin{tabular}{l ccccc | ccc | ccc}
    \toprule
    Source &       Daily Kos & HuffPost & CNN & WaPo & NYTimes & USA Today & AP & The Hill & WashTimes & Fox News & Breitbart \\
    Ideology & L & L & L & L & L & C & C&C & R & R & R\\
    \midrule
    \bnimgcap & 58k & --- & 300k & --- & 96k & --- & 370k & --- & 116k & 41k & 251k \\
    \bnaimg & 93k & 221k & 56k & 92k & 94k & 156k & 253k & 318k & 212k  & 303k  & 184k  \\
    \bottomrule
    \end{tabular}

    \caption{
    Number of image-text pairs in our newly-collected \textbf{pretraining} datasets, separated by news source. \bnaimg contains article text, while \bnimgcap contains captions.
    }
    \label{tab:img-data-summary}
    \vspace{-1mm}
\end{table*}

\subsection{Evaluation Datasets}

\paragraph{AllSides}
We extract a dataset of news articles and images from AllSides,
which associates \textit{stories} (e.g., of a particular event) with multiple \textit{articles} about that story but written by various news sources across a 5-point ideology scale (left, lean-left, center, lean-right, and right).
We crawl the AllSides website to obtain (story, article, source) tuples from 2012/06/01--2021/08/31, focusing on articles from the 25 news sources with the most number of articles in AllSides and spanning the complete range of ideology (see \Cref{sec:allsides-sources} for the complete list).
For each news article, we extract the article text and cover image from each article's corresponding news source's website, totaling 5,662 stories containing 12,471 articles.

\paragraph{Reddit}

\begin{figure}
    \vspace{-2mm}
    \centering
    \includegraphics[width=1.02\columnwidth]{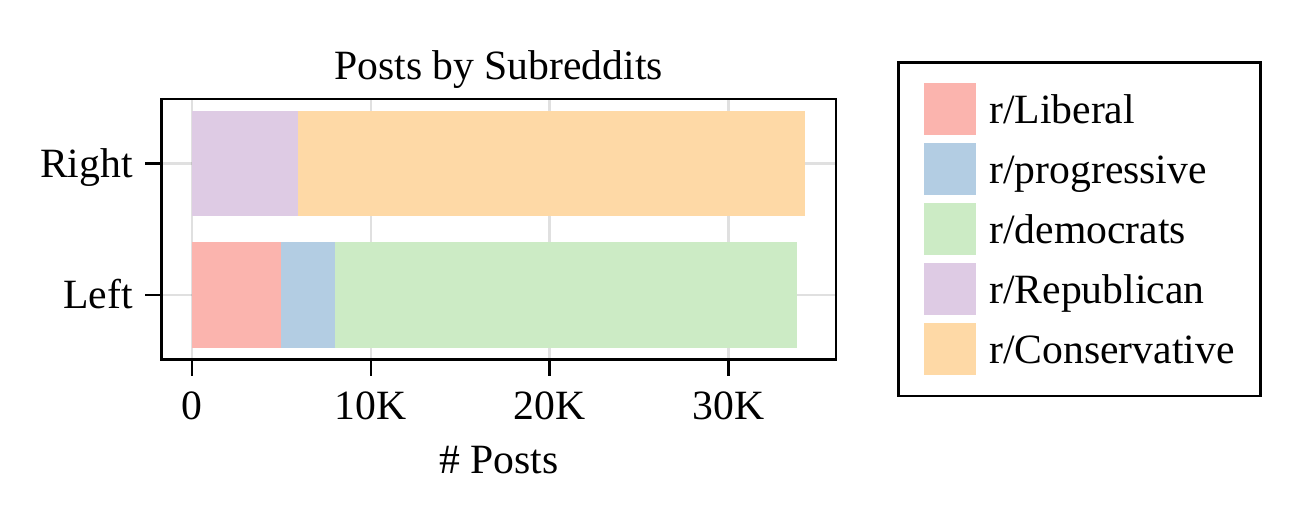}
    \vspace{-0.75cm}
    \caption{Proportion of posts with images from each political subreddit.}
    \vspace{-1mm}
    \label{fig:reddit-subs}
\end{figure}

We also collect a dataset of 357k Reddit posts with images from the past 10 years from five subreddits representing both the left-leaning (r/Liberal, r/democrats, r/progressive) and right-leaning (r/Conservative, and r/Republican) political stances, chosen for being among the largest and most active partisan subreddits. For each post, we keep the post title and the image itself, as long as the post was not removed (\~{}2,300 posts).
In order to avoid data leakage, we filter out all posts linking to images from the 11 news sources represented in \textsc{BigNews}, resulting in a set of 313k posts. In addition, because the number of posts from right-leaning subreddits overwhelms the number of left-leaning posts, we subsample the right-leaning posts, resulting in a balanced dataset of 65k posts with images, half from each political leaning. The Reddit dataset is summarized in \Cref{fig:reddit-subs}.
In contrast to news articles, Reddit imposes a 300-character limit on the post title.\footnote{We considered using the post's selftext (i.e. description), but only 101 of our collected posts contained any selftext. The majority of titles are under 100 characters (\~{}20 words).} Thus, this dataset and the Twitter data described below provide a good opportunity to examine how our models perform on \emph{short} texts, compared to the longer-form news articles.

\paragraph{Twitter}

\begin{figure}
    \centering
    \includegraphics[width=\columnwidth]{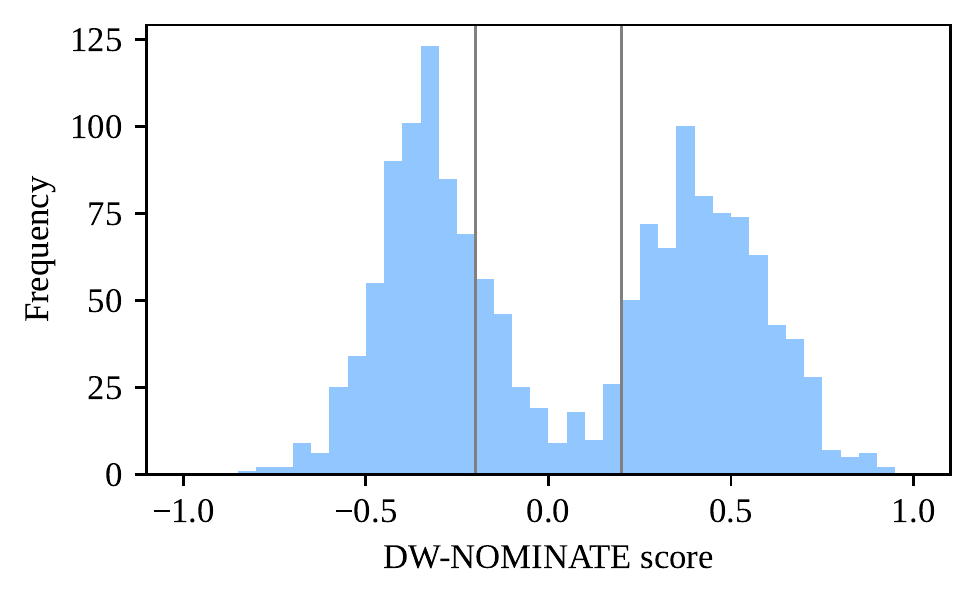}
    \caption{
    Histogram of the first DW-NOMINATE dimension in VoteView. Negative indicates left-leaning, while positive indicates right-leaning. Gray bars indicate the split points at -0.2 and 0.2 that separate the left, center, and right ideologies.}
    \label{fig:voteview-ideology}
\end{figure}

We additionally collect a dataset of 2.1M political tweets from Twitter from the past 10 years
using the Twitter Decahose stream, selecting tweets by political figures included in a list of 9,981 US politicians and their Twitter handles \citep{panda2020nivaduck}.
In contrast to AllSides, Twitter does not explicitly annotate discrete ideologies.
Thus, we label tweets with their author's ideology, identified based on their DW-NOMINATE\footnote{The DW-NOMINATE scores are obtained from \url{VoteView.com}.} dimension \citep{boche2018new}, a measure of a politician's voting history: a positive number indicates a conservative leaning (e.g. Donald Trump, 0.403), while a negative number indicates a liberal leaning (e.g. Barack Obama, -0.343).
We partition politicians into left, center, and right ideologies, containing those whose ideology score is less than -0.2, between -0.2 and 0.2, and above 0.2, respectively.
The distribution of these scores is shown in \Cref{fig:voteview-ideology}.
Finally, we discard tweets without images, leaving 57,093 tweets from 1,422 politicians as our final evaluation dataset. More details are summarized in \Cref{tab:twitter-post-stats}.

\begin{table}[t]
    \centering
    \small
    \begin{tabular}{lrrrrr}
        \toprule
        Ideology & Users & Tweets & Mean & Median & Std \\
        \midrule
        Left   & 523 & 26,362 & 50.4 & 37 & 48.7 \\
        Center & 137 &  6,963 & 50.8 & 34 & 54.1 \\
        Right  & 628 & 23,768 & 37.9 & 25 & 41.6 \\
        \bottomrule
    \end{tabular}
    \caption{Total number of politician users and Twitter posts in our dataset, with associated statistics per user (last three columns).}
    \label{tab:twitter-post-stats}
\end{table}

\subsection{Characterization and Analysis of Datasets}
\label{sec:image-analysis}

To motivate different model and pretraining variants described in the next section, we analyze the content of images and text in our newly-collected AllSides dataset using both automatic and manual means.

\paragraph{Automatic Annotation of Images}
The majority of images contain political figures; we wish to identify these figures\footnote{Figure, face, and person are synonymous in this work. Faces are used in the facial detection process by DeepFace, but in some cases, faces are not visible and are thus not identifiable.}
and some salient aspects that may be relevant to predicting the ideology of the article. We employ DeepFace \citep{taigman2014deepface}, a state-of-the-art facial recognition framework. Given an input image, DeepFace identifies faces and matches them to a set of reference images; we construct a set of 10 reference images for 722 political figures using a combination of entity linking and manual heuristics, detailed in \Cref{sec:deepface-construction}. We also employ DeepFace to detect
gender (male/female), race (Asian, Black, Indian, Latino, Middle Eastern, or White), and emotion (neutral, angry, fear, sad, disgust, happy, surprise) in AllSides images.\footnote{These categories are those supported by DeepFace, not specifically chosen by us.}

\paragraph{Pitfalls of Facial Recognition}
While using DeepFace, we encountered a few pitfalls. First, DeepFace is often unable to recognize faces that are small, blurred, or in side profile. This corroborates existing work showing that reduced quality of faces is detrimental to the detection of faces and emotions \citep{jain2010fddb,yang2016wider,yang2021benchmarking}. Second, we noticed frequent mistakes with a few high-profile figures. For example, DeepFace often classifies Barack Obama and Eric Holder as Hispanic or Middle Eastern, and Donald Trump as Asian, showing that DeepFace can be faulty even for famous people with lots of training images.

\paragraph{Manual Annotation of Images}
No facial recognition tool is perfect, and aspects of images that could be relevant for ideology, such as main figures or the presence of certain objects, are not captured by DeepFace. Therefore, we manually annotate 400 random images from AllSides. For each image, we identify the \textbf{number of people} in the image (1-5, or ``group'' if there are 6 or more people). We identify the \textbf{main figure(s)} in the image. For each main figure, we identify their \textbf{name} (if a known political figure), \textbf{gender}, \textbf{race}, and \textbf{emotion} (Positive, Negative, or Neutral).\footnote{We only annotate three categories of emotion, because we found it hard to distinguish between the fine-grained negative emotions detected by DeepFace.}
If the figure is of mixed race (e.g., Barack Obama) or if the figure is unknown (i.e., not easily identifiable after examining the article's text and searching Google), we label their most salient race.

\begin{figure}[t]
    \centering
    \begin{subfigure}[b]{0.52\columnwidth}
    \includegraphics[width=\columnwidth]{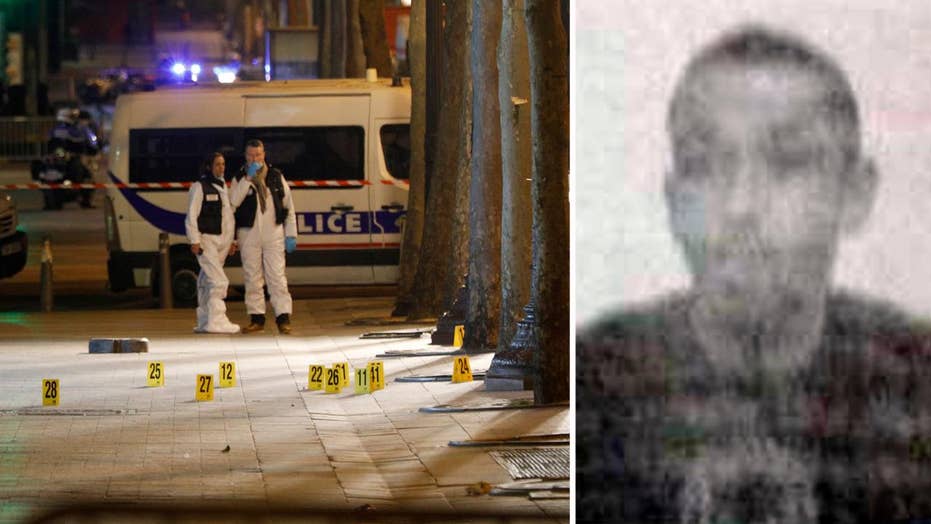}
    \caption{Collage}
    \label{fig:multi-image}
    \end{subfigure}
    ~
    \begin{subfigure}[b]{0.44\columnwidth}
    \includegraphics[width=\columnwidth]{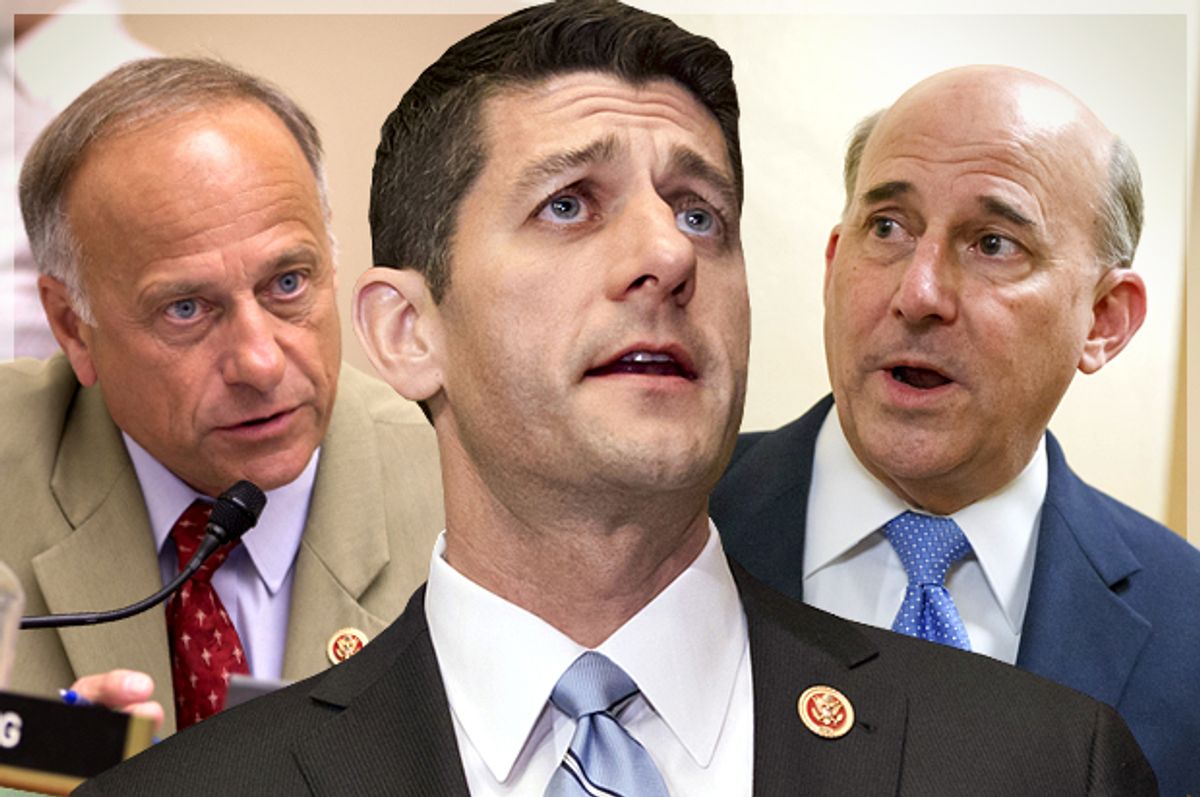}
    \caption{Composite Image}
    \label{fig:composite-image}
    \end{subfigure}

    \caption{Collages are composed of separate images arranged adjacently, while a composite image is composed of partial images edited together. Collages are often used to tell a sequential story, while composites show a connection between different people.
    }
    \label{fig:multi-vs-composite}
\end{figure}

We also identify any \textbf{salient aspects} of the image that help convey the image's message. This may include the presence of certain objects (e.g., guns, flags), activities (e.g., protests), or text in the image. We also annotate special \textbf{image classes}: whether the image is an invalid/missing image, a news source banner, a cartoon drawing,
a collage, or a composite image. The difference between the latter two is explained in \Cref{fig:multi-vs-composite}.

\begin{table}[t]
    \centering
    \scriptsize
    \scalebox{0.99}{
    \begin{tabular}{lrrrrr}
        \toprule
         & Left & Lean Left & Center & Lean Right & Right \\
        \midrule

        No Face   & 18\% & 16\% & 19\% & 12\% & 15\% \\
        1 Face    & 41\% & 38\% &  4\% & 49\% & 46\% \\
        2 Faces   & 14\% & 14\% & 14\% & 14\% & 16\% \\
        3 Faces   &  5\% &  9\% &  7\% &  8\% &  7\% \\
        4 Faces   &  6\% &  6\% &  4\% &  4\% &  4\% \\
        5+ Faces  & 16\% & 18\% & 16\% & 12\% &  12\% \\
        \midrule
        Mean \# faces & 1.89 & 2.04 & 1.87 & 1.80 & 1.74 \\
        Total \# Images & 665 & 2152 & 1142 & 923 & 2058 \\
        \bottomrule
    \end{tabular}
    }

    \caption{Percentage of images containing faces in AllSides, analyzed by DeepFace. On average, left-leaning images use slightly more figures than right-leaning images.}
    \label{tab:image-analysis}
\end{table}

\begin{table}[t]
    \centering
    \scriptsize
    \begin{tabular}{lrrrrr}
        \toprule
        & Left & Lean Left & Center & Lean Right & Right \\
        \midrule
        Regular   & 85\% & 93\% & 100\% & 95\% & 85\% \\
        Removed   &  7\% &  0\% &   0\% &  0\% &  0\% \\
        Banner    &  2\% &  0\% &   0\% &  2\% &  2\% \\
        Cartoon   &  0\% &  0\% &   0\% &  2\% &  0\% \\
        Collage   &  0\% &  6\% &   0\% &  0\% & 12\% \\
        Composite &  5\% &  1\% &   0\% &  0\% &  1\% \\
        \bottomrule
    \end{tabular}
    \caption{Type of images in a random sample of 400 images from AllSides. While most images are ordinary, a small percentage fall under special cases. Notably, a large fraction of Right images are collages (i.e., 12\%), indicating a common strategy by right-leaning media.
    }
    \label{tab:image-class}
\end{table}

\paragraph{Analysis}
We present annotator agreement between DeepFace and humans in \Cref{sec:deepface-agreement}. In this section, we focus on drawing insights from the analysis of the images.

We first examine the number of figures in the image (\Cref{tab:image-analysis}). We find that images from liberal sources on average contain more figures than images from conservative sources. Specifically, a higher percentage of left-leaning and center images contained 5 or more faces. Within the 5+ faces category, a large fraction are unknown figures (i.e., not well-known politicians), though these images may contain more notable politicians (e.g., Trump at a podium surrounded by supporters). The distribution of the number of figures in these images may reflect liberals' focus on equality as a group in comparison to conservatives' focus on self-reliance as part of their political identity, as revealed by prior work~\citep{hanson2019individual}.

We also examine the distribution of face occurrence by topic of the article.
We find images about topics such as civil rights, labor, and holidays have the most figures on average, while topics such as national defense, FBI, and criminal justice have relatively fewer people. This is simply a natural reflection of the nature of the topic. Within a topic, the distribution largely follows the \textit{liberal images have more figures} rule. For example, in the gun control topic, lean-left images contain on average 4.9 people, while lean-right images contain on average 2.0 people.

Roughly 12--19\% of images contain no face. We find that the majority of these images contain inanimate subjects mentioned in the news articles, which are about e.g. an oil tanker that caught on fire, or rubble from an earthquake, rather than about a specific political figure. Around 13\% of these no-face pictures contain well-known government buildings including the White House, the Capitol building, and the Supreme Court building. One explanation is that these images represent the three branches of government in the US and are thus a form of metonymy, e.g. the White House can refer to not only the president but also the country as a whole.
However, future work is needed to understand \textit{why} reporters would select, e.g., an image of the White House instead of an image of the president.

We also investigate types of images in \Cref{tab:image-class}. Most are ordinary images, but
we find that over 12\% of images from Right sources are collages, which are often arranged in the form of a narrative. For example, in \Cref{fig:multi-image}, from an article describing a bombing in France, the first image contains the police scene, while the second image is the suspect.  Existing work has demonstrated links between the usage of collages and the dissemination of misinformation \citep{krafft2020disinformation}.
Composite images make up over 5\% of Left images, and we observe that these images are often used to indicate confrontation between the figures in the image, such as two contestants in an election, or policymakers who disagree on an issue.

The four most frequent figures in images are Donald Trump, Barack Obama, Hillary Clinton, and Joe Biden (see \Cref{tab:image-figures-analysis} for details). We find a trend that articles from a particular ideology tend to have more images of the opposing figure (e.g., right-leaning media contains more images of Obama). This is likely a result of attack politics \citep{haynes1998attack,theilmann1998campaign}, where politicians attack their opponent instead of bolstering their own position, especially when campaigning. This type of negative campaigning has been shown to be employed more by Republicans \citep{lau2001negative}.

\begin{figure}
    \centering
    \vspace{-1mm}
    \includegraphics[width=0.9\linewidth]{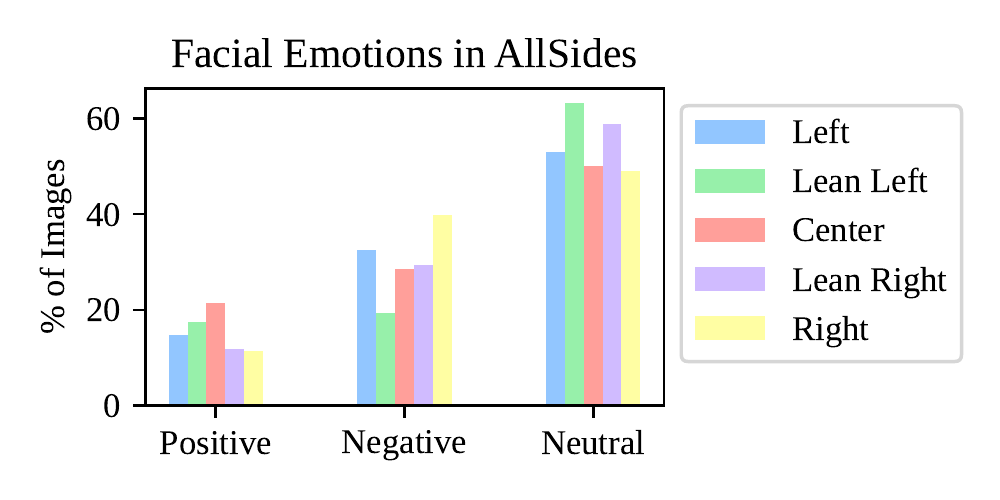}
    \vspace{-0.25cm}
    \caption{Facial emotions stratified by ideology from human annotations of AllSides. The majority of emotions are negative or neutral, rather than positive. Notice that Left and Right (i.e., the media labeled as more extreme ideologies) have a much higher proportion of negative faces than Lean Left and Lean Right.}
    \vspace{0mm}
    \label{tab:emotion}
\end{figure}

Lastly, we analyze the emotion of the figures in AllSides (\Cref{tab:emotion}). Across all ideologies, the majority of faces have negative or neutral emotion. Consumers actually prefer negative news over positive news \citep{trussler2014consumer}, and negative images in news are more memorable \citep{newhagen1992evening,newhagen1998tv}. Specifically for facial expressions, liberal and conservative news sources have differences in portrayals of Donald Trump \citep{boxell2021slanted}. Angry facial emotion primes also tend to increase trust in negative news messages \citep{ravaja2014suboptimal}. This may explain why more extreme Left and Right news sources, which are more likely to contain less credible news \citep{allen2020evaluating}, have a higher rate of negative emotion faces than Lean-Left and Lean-Right.

%% file: sections/04-models.tex
\section{Models}

Armed with new diverse datasets of articles and images, we now propose several models, input encoding strategies, and pretraining regimens to tackle the challenges of multimodal ideology prediction.

\begin{figure}
    \centering
    \vspace{-1mm}
    \includegraphics[width=\columnwidth]{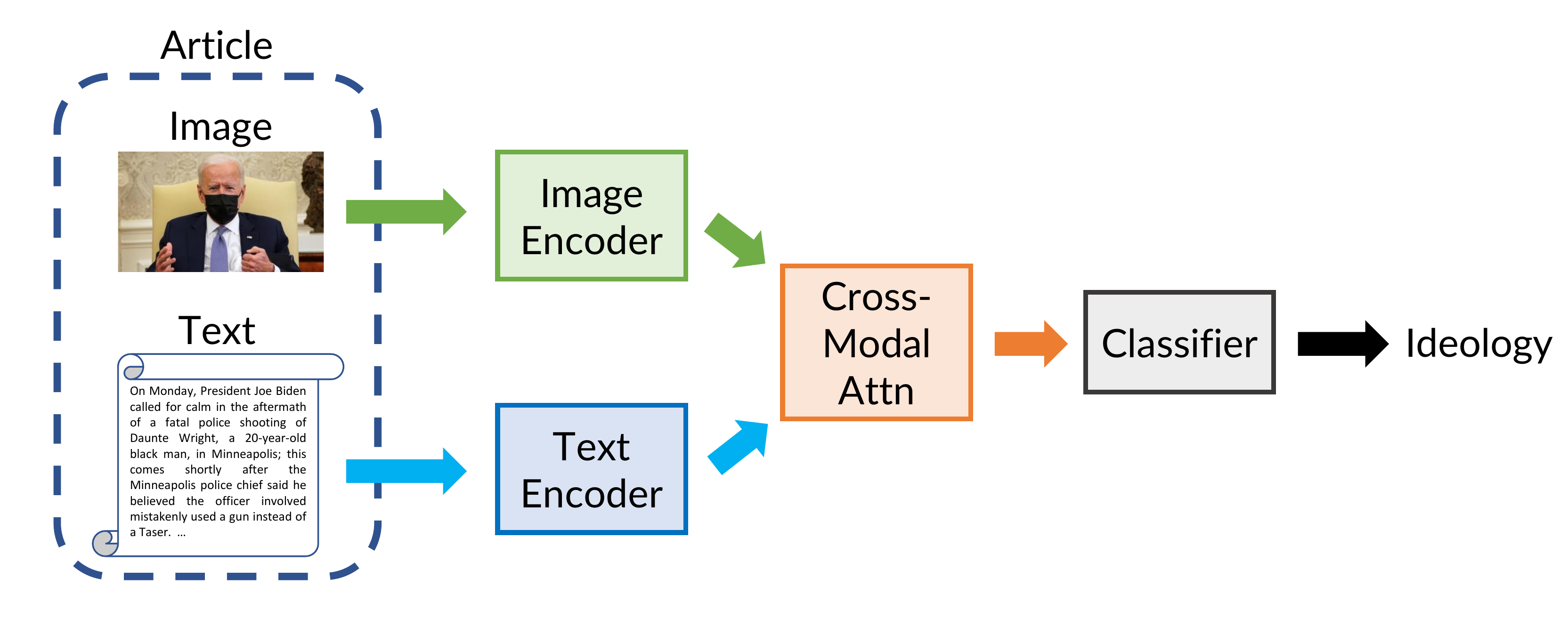}
    \caption{High-level structure of the late-fusion model architecture. The representations of the image and the text are separately computed, then combined before being passed to a classification layer.}
    \label{fig:model-diagram}
\end{figure}

\subsection{Text-Only and Image-Only Models}

We first experiment with text-only models, including RoBERTa \citep{liu2019roberta} and POLITICS \citep{liu2022politics}, a RoBERTa model further pretrained with a political ideology objective and thus specialized for ideology prediction and stance detection.

For image-only models, we use Swin Transformer \citep{hu2019local,liu2021Swin,liu2021swinv2,xie2021simmim}, a general-purpose hierarchical Transformer-based model that computes representations of images using shifted windows and has obtained strong or state-of-the-art performance on several image processing tasks.
Because of our focus on faces, we experiment with several face-aware image preprocessing methods before encoding the images. These methods are described in \Cref{app:image-preprocessing} but were ultimately not successful. Thus, we use the images unchanged.

\subsection{Multimodal Models}

\paragraph{Early Fusion} Also known as single-stream, an early fusion model takes the joint sequence of text and images as input and merges both modalities to obtain a single representation. We experiment with VisualBERT \citep{li2019visualbert} and ViLT \citep{kim2021vilt}, two Transformer-based models that have demonstrated strong performance on a series of vision-and-language downstream tasks such as VQAv2 \cite{balanced_vqa_v2}, NLVR2 \cite{DBLP:conf/acl/SuhrZZZBA19}, and Flickr30K \cite{plummer2015flickr30k}.
VisualBERT concatenates words and image segments identified by an object detector, with an additional embedding indicating the input modality.
Instead of using object detectors, we feed in faces detected by DeepFace, as we consider political figures more relevant to the ideology prediction task.
ViLT is a similar architecture, but with separate positional embeddings for the text and image inputs and does not use an object detector. We use the pretrained weights released publicly by the authors.

\paragraph{Late Fusion} Also known as dual-stream, two models separately encode each modality; then the two representations are joined into a single representation. This is in contrast to early fusion, where a single encoder processes the image and text jointly. We use RoBERTa to encode the text, and Swin Transformer to encode images.

We evaluate several representation joining mechanisms: concatenation, Hadamard product $\bigodot$, gated fusion \citep{wu-etal-2021-good}, and cross-modal attention \citep[LXMERT;][]{tan-bansal-2019-lxmert}.
Gated fusion combines the two representations by learning a gate vector $\lambda$ so that the combined representation is $\mathbf{h} = \mathbf{h}_\textrm{text} + \lambda \bigodot \mathbf{h}_\textrm{img}$.
For cross-modal attention, \cite{DBLP:journals/tacl/HendricksMSAN21} has comprehensively analyzed different types of attention mechanisms and found that the \textit{coattention} scheme (given queries from one modality, e.g., image, keys and values can be taken only from the other modality, e.g., language) has the best performance. Therefore, we use the \textit{co-attention} scheme for our cross-modal attention module; our implementation largely resembles the cross-modal attention module in LXMERT, with the number of cross-modality layers $\mathbf{N_X}$ increased from 5 to 6.

\subsection{Continued Pretraining to Inject Knowledge of Ideology}
\label{sec:pretraining}

Recent work has shown that continuing to train a pretrained model on domain-specific data or on an auxiliary task can improve the model's performance on the target task \citep{beltagy-etal-2019-scibert,gururangan-etal-2020-dont,lee2020biobert}. In this vein, we aim to improve our ideology prediction model by performing continued pretraining with relevant objectives and auxiliary data.

For pretraining the image encoder, we experiment with an \textbf{InfoNCE loss} \citep{sohn2016improved,van2018representation,radford2021learning}, a contrastive loss computed within each batch, where the image and text encoders are trained to maximize the cosine similarity of the image and text embeddings of the $n$ correct pairs in the batch, while minimizing the cosine similarity of the embeddings of the $n^2 - n$ incorrect pairings.
We use this loss with images and their captions, with the hypothesis that supervision from captions will allow the image encoder to develop a more robust representation and potentially learn features of the image that are present in the caption.

We also experiment with a \textbf{bidirectional captioning loss} \citep[VirTex;][]{desai2021virtex}, in which the image embedding is passed to an image captioning Transformer head, which generates a corresponding caption token by token in both the left-to-right and right-to-left directions.

Finally, we propose a novel \textbf{triplet margin loss} on triplets of news (anchor, positive, negative), where the positive pair shares the same ideology with the anchor, while the negative image has a different ideology than the anchor. Mathematically,
$
    \mathcal{L} = \sum_ {\mathbf{t}\in T} \left[ ||\mathbf{t}^{(a)} - \mathbf{t}^{(p)}||_2
                - ||\mathbf{t}^{(a)} - \mathbf{t}^{(n)}||_2 + \alpha \right]_+
$,
where $T$ is the set of news triplets in the training set;
$\mathbf{t}^{(a)}$, $\mathbf{t}^{(p)}$, and $\mathbf{t}^{(n)}$ are the joint representations of text and image (concatenated and passed through a linear transformation) of the anchor, positive, and negative news in triplet $\mathbf{t}$.
$\alpha$ is a bias term;
and $\left[\cdot\right]_+$ is the ramp function $\mathrm{max}(\cdot, 0)$. This is inspired by the triplet loss used in FaceNet \citep{schroff2015facenet} and is similar to the triplet ideology loss proposed by \citep{liu2022politics}, who pretrain a text-only ideology prediction model with this loss. We apply this loss to pretrain the image encoder, text encoder, and embedding combination components of our model.

%% file: sections/05-experiments.tex
\section{Experiments}

We first perform preliminary experiments comparing single-modality models to determine whether the inclusion of images helps ideology prediction. Then, we evaluate multimodal experimental setups, exhaustively selecting an image encoder, embedding combination methods, and pretraining methods; performing the continued pretraining; then finetuning the entire model on the task of ideology prediction.

\paragraph{Implementation Details}
We implement the new models in PyTorch, importing existing models from their authors' respective GitHub pages.
All models were trained for a maximum of 20 epochs with early-stopping patience of 4. For detailed hyperparameters for pretraining and finetuning and other specific implementation details, please refer to our code on the project page.

%% file: sections/06-results.tex
\section{Results and Analysis}

\setlength\dashlinedash{2pt}
\setlength\dashlinegap{1.5pt}
\setlength\arrayrulewidth{0.1pt}

\begin{table}
    \centering
    \tiny
    \begin{tabular}{lllcc}
        \toprule
        \textbf{Text Enc} & \textbf{Image Enc} & \textbf{Image Prep} & \textbf{Acc.} & \textbf{Macro} $F_1$ \\
        \midrule
        RoBERTa& --- & ---&
        85.14   ±   0.50 & 84.41   ±   0.58 \\ \hdashline
        --- & Swin-T & Full Image & 48.94  ±  0.68 & 50.49  ±  0.70 \\
        --- & Swin-S & Full Image & \textbf{49.33  ±  0.41} & \textbf{50.63  ±  0.39}\\
        --- & Swin-B & Full Image & 48.26  ±  0.89 & 49.52  ±  0.93 \\
        --- & Swin-S & Only Faces &  25.31  ±  0.00 & 8.08  ±  0.00 \\
        \bottomrule
    \end{tabular}
    \caption{Experiments with single modality models (no pretraining) on 5-way prediction on AllSides. All results are averaged over 5 runs. RoBERTa is already a strong baseline, showing that an article's text is sufficient in many cases for predicting ideology. However, the image-only Swin models perform quite poorly; in many cases, it is hard to infer ideology solely from images.
    }

    \label{tab:exp-unimodal}
\end{table}

\begin{table}[t]
    \centering
    \scriptsize
    \begin{tabular}{clcc}
        \toprule
        \textbf{Category}      & \textbf{Model}     &\textbf{ Acc.} & \textbf{Macro $F_1$} \\
        \midrule
        \multirow{2}{*}{Early Fusion}   & VisualBERT  & 78.45 ± 0.69 & 75.34 ± 0.67 \\
                      & ViLT        & 78.39 ± 1.24 & 76.22 ± 1.43\\\hdashline
        \multirow{5}{*}{Late Fusion} & \multicolumn{2}{l}{RoBERTa+Swin-S}  \\
            & \quad Concat.           & 82.39 ± 0.59 & 79.82 ± 1.02	\\
            & \quad Hadamard Prod.    & 85.14 ± 0.74 & 82.62 ± 1.15 \\
            & \quad Gated Fusion     & 82.77 ± 1.24 & 80.71 ± 1.20  \\
            & \quad Cross-modal Attn. & \textbf{86.88 ± 0.38} & \textbf{85.47 ± 0.41} \\
        \bottomrule
    \end{tabular}
    \caption{Multimodal results on AllSides without pretraining. Early fusion models perform worse than text-only baselines (found in \Cref{tab:exp-unimodal}). In late fusion models, the embedding joining methods show no significant difference. However, the best performing model with no pretraining is the late fusion RoBERTa+Swin model with cross-modal attention.}
    \label{tab:multimodal}
\end{table}

\begin{table*}[t]
    \centering
    \scriptsize
    \scalebox{0.88}{
    \begin{tabular}{lllccccccc}
        \toprule
           \multicolumn{3}{c}{\textbf{Pre-training Component \& Objective}} & \multicolumn{6}{c}{ \multirow{1}{*}{ \textbf{Acc.} }} & {\textbf{Macro F1}} \\
           \textbf{Text Enc.}  &  \textbf{Image Enc.} & \textbf{Cross-modal Attn. }  & Overall Acc. & Left & Lean Left & Center & Lean Right & Right & {Overall F1} \\
        \midrule
            \xmark& \xmark& \xmark  & 85.47 ± 0.41  &69.10 ± 3.34 & 88.03 ± 0.85 & 91.41 ± 0.83 & 88.15 ± 1.31 & 76.40 ± 2.38 & {82.62 ± 1.15} \\
               \cmark (POLITICS) & \xmark& \xmark  & 86.80 ± 0.72 & \textbf{96.42 ± 0.96} & \textbf{92.42 ± 1.95} & 90.36 ± 0.93 & 81.80 ± 1.45 & 72.58 ± 3.43 & {86.39 ± 0.72} \\
               \xmark & \cmark (InfoNCE) & \xmark & 86.40 ± 0.74 & 88.35 ± 1.24 & 90.09 ± 0.50 & 89.96 ± 0.88 & 82.37 ± 1.04 & 81.24 ± 1.84 & {86.12 ± 0.73} \\
           \xmark & \cmark (VirTex-style) & \xmark &87.60 ± 0.47 & 88.72 ± 1.89 & 89.77 ± 0.73 & 89.87 ± 0.91 & 85.58 ± 0.74 & 84.06 ± 0.84 & {87.84 ± 0.45} \\
         \xmark & \xmark & \cmark (Triplet Margin)   &87.86 ± 0.93 & 87.99 ± 0.62 & 90.88 ± 0.76 & 90.67 ± 1.68 & 86.19 ± 0.61 & 83.59 ± 2.16 & {87.45 ± 0.87} \\
                           \cmark (POLITICS) & \cmark (VirTex-style) & \xmark & 88.49 ± 0.58 &94.03 ± 1.30 &  88.87 ± 2.65 & \textbf{92.30 ± 2.73} &\textbf{88.62 ± 2.17} & 78.52 ± 2.22 & {88.26 ± 0.88}
\\
                           \xmark & \cmark (VirTex-style) &  \cmark (Triplet Margin) &88.18 ± 0.56 &87.80 ± 0.99 &  90.32 ± 0.92 & 91.83 ± 1.11 & 85.98 ± 0.76 & \textbf{84.97 ± 0.94} & {88.16 ± 0.59} \\
              \cmark (POLITICS) & \cmark (VirTex-style) & \cmark (Triplet Margin)   &\textbf{88.98 ± 0.65} & 91.04 ± 1.50 & 91.08 ± 0.59 & 91.36 ± 0.84 &85.82 ± 0.55 & 84.90 ± 0.57 & {\textbf{88.64 ± 0.68}}   \\
        \midrule
        \bottomrule
    \end{tabular}
    }
    \caption{Pretraining ablation experiments on AllSides. The base model is RoBERTa + Swin-S. We report mean and standard deviation over five runs. The base model performs poorly on Left. Adding pretraining substantially improves performance overall, especially on articles reported by the Right-leaning media.
    }
    \label{tab:pre-train-allsides}
\end{table*}

\begin{table*}[t]
    \centering
    \scriptsize
    \begin{tabular}{llcclcc}
        \toprule
                     &     &\multicolumn{2}{c}{Reddit} & &\multicolumn{2}{c}{Twitter (2-way, center filtered out)} \\
        \textbf{Category}      & \textbf{Model}     &\textbf{ Acc.} & \textbf{Macro $F_1$} & &\textbf{ Acc.} & \textbf{Macro $F_1$}\\
        \midrule
        \multirow{2}{*}{Text-only}   & RoBERTa        &76.82 ± 0.32 &76.81 ± 0.32 & & 77.70 ± 0.43 &77.53 ± 0.45\\
        & POLITICS     &76.68 ± 1.57  &76.42 ± 1.81 &   & 78.80 ± 0.15 & 78.78 ± 0.14 \\
        Vision-only & Swin-S          &  70.90 ± 0.55 &70.87 ± 0.58  & & 62.53 ± 0.85 & 62.49 ± 0.84\\
        \hdashline
        \multirow{8}{*}{Late Fusion}
        & RoBERTa+Swin-S+Cross-modal Attn.      &   & 	 \\
        &        \quad No Further Pre-training &77.79 ± 0.17 &77.72 ± 0.15 & & 78.50 ± 0.06 &62.97 ± 0.66 \\
        & \quad VirTex-style+Triplet Margin  &  80.82 ± 1.05 &80.78 ± 1.09  & & 79.49 ± 0.44 &78.46 ± 0.06 \\
         & POLITICS+Swin-S+Cross-modal Attn.     & & \\
          &        \quad No Further Pre-training &79.06 ± 0.14 &79.02 ± 0.14  & & 78.82 ± 0.69 &78.80 ± 0.68\\
        &        \quad VirTex-style+Triplet Margin  &  \textbf{81.72 ± 0.87} &\textbf{81.69 ± 0.69} & &\textbf{79.85 ± 0.18}  &\textbf{79.82 ± 0.20} \\
        \bottomrule
    \end{tabular}
    \caption{Results on Reddit \& Twitter datasets, showing mean and standard deviation over five runs. Due to domain mismatch, performance on Reddit and Twitter is worse than on AllSides. However, the addition of pretraining improves overall performance. For a detailed breakdown by ideology, see \Cref{tab:reddit-exp,tab:twitter-exp}.}
    \label{tab:reddit-twitter-exp}
\end{table*}

\paragraph{Text-Only and Image-Only Models}

We first present unimodal experiments on AllSides in \Cref{tab:exp-unimodal}.
We find that the vision-only model performs significantly worse than the text-only baseline, indicating that an image alone is inadequate for predicting ideology.
Surprisingly, Swin-Small slightly outperforms Swin-Base (which is larger in size), though the difference is not substantial.
Thus, we decided to use Swin-Small (Swin-S) as the image encoder backbone for our multimodal models due to its performance and size.
These baseline results motivate the premise of \textit{multimodal} ideology prediction, in which we use images as additional signal to augment the text.

\paragraph{Multimodal Models without Pretraining}

Next, we present multimodal model results without pretraining in \Cref{tab:multimodal}.
We find that early fusion models cannot outperform the text-only baselines, in contrast to late fusion models, indicating that the combination of text and image is beneficial for ideology prediction, but the choice of architecture is important. The late fusion architecture with cross-modal attention performs the best, and thus we take this model as our starting point for the rest of the experiments in this paper.

\paragraph{Multimodal Models with Pretraining}
We then exhaustively experiment with combinations of pretraining objectives for each component of the model (language encoder, image encoder, and cross-modal attention) for ablation and analysis. Results on the AllSides evaluation set are shown in \Cref{tab:pre-train-allsides}.

First, we find that replacing RoBERTa with the pretrained POLITICS model already gains a 1\% improvement to the overall model. By pretraining on similar domain text, the model is able to generate better text representations. \citet{liu2022politics} find that the POLITICS objective allows the model to perform much better on left-leaning articles, which have higher perplexity (i.e. the language is more diverse). In a multimodal setup, we find that this text encoder helps the multimodal model improve on right-leaning input data more than left-leaning, indicating that the inclusion of images helps the classification of right-leaning ideology.

For pretraining the image encoder, our experiments show that the VirTex-style bidirectional captioning loss performs better than the InfoNCE contrastive learning objective. This pretraining method allows the model to better capture the similarities between the image and its associated text. Text may also provide a more semantically dense signal than contrastive approaches \citep{desai2021virtex}, thus leading to better performance.

For pretraining the cross-modal attention, we find that our proposed Triplet Margin Loss objective, which optimizes all three components (image encoder, text encoder, cross-modal attention) of the entire model, improves over no pretraining.

Overall, ablation experiments show that the best combination (RoBERTa pretrained with POLITICS loss, and Swin pretrained with VirTex loss on the image encoder) contribute around 1 percentage point and 2 percentage points respectively (\Cref{tab:pre-train-allsides}). Further combining them with the triplet margin loss leads to the best-performing model, with more than 3 percentage points over the baseline late-fusion model.

\paragraph{Twitter \& Reddit}
Finally, we evaluate our models on the Reddit and Twitter datasets to get a more comprehensive perspective of the model's ideology prediction ability on different domains. Results are presented in \Cref{tab:reddit-twitter-exp}. Overall, performance is substantially lower than on AllSides, because of domain mismatch: Reddit posts and tweets are not usually written in the long, formal language of news articles.
However, the improvements over a text-only baseline are more substantial than on AllSides, where the long text already contains enough information for predicting ideology. We also find that for Twitter, which was split into left, center, and right ideologies, the models perform poorly on Center tweets, probably due to high dataset imbalance (\Cref{tab:twitter-post-stats}), though the addition of images greatly improved over text-only models.

%% file: sections/07-conclusion.tex
\section{Conclusion}

This paper introduces the task of fine-grained multimodal ideology prediction, where a model predicts one of five ideologies leanings given a pair of text and an image. We collect five new large-scale datasets of political images and present an in-depth characterization of these images, examining aspects such as facial features across ideologies. We experiment with a combination of state-of-the-art multimodal architectures and pretraining schemes, including a newly-proposed triplet margin loss objectives. Along with the release of our datasets, our experimental findings will inform the selection of models and training objectives in future work and spur future research in politics, ideology prediction, and other multimodal tasks.

%% file: sections/08-ethics.tex
\section*{Ethical Considerations}

\subsection*{Dataset Collection}
All images were collected in a manner consistent with the terms of use of the original sources. The articles and images from AllSides, Reddit, Twitter, and the 11 news sources are copyrighted by their respective sources. We consulted Section 107\footnote{\url{https://www.copyright.gov/title17/92chap1.html\#107}.} of the U.S. Copyright Act and ensured that our collection action fell under the fair use category. As we are not the copyright holders, we do not release the images that we collected. Rather, we provide code for analyzing these datasets for those who may already have the datasets.
In addition, to discourage the misuse of the data, we warn users about potential misuse and any ethical concerns that could raise from improperly dealing with the data.

\subsection*{Facial Recognition}
We use DeepFace to perform facial recognition as well as attribute recognition for gender, race, and emotion. However, in this work, we only use DeepFace to perform analysis (rather than prediction) on our new datasets, and we compare DeepFace's analyses with human annotations. The categories for gender, race, and emotion are defined by DeepFace. Though some may question these categories, it is not within the scope of this paper to argue for or against them.

As mentioned in \Cref{sec:image-analysis}, DeepFace often classifies Barack Obama and Eric Holder as Hispanic or Middle Eastern, and Donald Trump as Asian. This is likely due to models learning that dark skin or squinty eyes, respectively, are important features predictive of race. As researchers, we must be aware of these biases in the models and be careful not to reinforce racial stereotypes due to models' predictions. We do not explicitly use race, gender, or ethnicity as features in our prediction model. Moreover, we call on all researchers to deal with automatic facial recognition tools like DeepFace carefully and take all possible biases into consideration.

\subsection*{Ideology}
In our work, we have made several assumptions about ideology, namely that the ideology of a news source, Reddit subreddit, and Tweet author is consistent. This may not always be true; a left-leaning post may appear in \texttt{r/Conservative}, or a politician on Twitter may be a moderate whose tweets reflect liberal and conservative stances on different issues. However, these are relatively rare cases, and we will warn all potential users about such cases.

\subsection*{Model}
\paragraph{Intended Use}
The use case we have described for our multimodal ideology prediction model is to educate users about ideology bias in media of various genres with both texts and images.

\paragraph{Failure mode}
The failure mode referred to a case where our model fails to predict the correct ideology of a piece of media work with both text and image. While we showed that these models have high accuracy, the models are not 100\% perfect. End users of our model must not take model predictions as fact. We encourage end users to consult machine learning experts as well as political scientists when using our models.

%% file: sections/09-limitations.tex
\section*{Limitations}

\paragraph{Facial Recognition} As our analysis has shown, DeepFace is not 100\% reliable as an automatic annotation tool. To confidently use DeepFace as an analysis tool, manual annotation (which we have done in the paper) is necessary but time-consuming, requiring human labor.

\paragraph{Compute Resources - GPUs} Due to the scale of the data (summarized in Table \ref{tab:img-data-summary}) and the size of the model (summarized in Table \ref{tab:param}), pretraining is extremely computationally expensive and requires large GPU resources. Our experiments are performed using 2 NVIDIA RTX A6000 and 2 Quadro RTX 8000 GPUs. Batch sizes are chosen to meet hardware constraints and we pretrain the models for 2500 steps. The pretraining on \bnimgcap and \bnaimg takes approximately 3 and 4 days, respectively.

\begin{table}[h]
    \centering
    \small
    \begin{tabular}{ll}
        \toprule
        \textbf{Model}     &\textbf{\# parameters} \\
        \midrule
        RoBERTa / POLITICS       &125M\\
        \hdashline
         Swin-T          &  29M\\
                  Swin-S          &  50M\\
                          Swin-B          &  88M\\
                          \hdashline
        RoBERTa+Swin-S & 175M \\
        RoBERTa+Swin-S+Cross-modal Attn. & 273M\\
        \bottomrule
    \end{tabular}
    \caption{Number of parameters in each model.}
    \vspace{-2mm}
    \label{tab:param}
\end{table}

%% file: sections/A-appendix.tex
\newpage
\section{Details of Image Collection}
\label{sec:image-collection}

This section contains details of the collection of images and their corresponding captions from news articles. Images were obtained by searching for \texttt{<img>} tags within the article's HTML. The captions were obtained as follows. We started with examining the \texttt{alt} attribute of the \texttt{<img>} tags, which often contain the caption. For Associated Press, we identified captions in a different part of the HTML and extracted them using regex matching. To ensure that captions adequately describe their images and prevent data spillage, we use regex matching to remove portions of the captions containing names of photographers and the news source name, such as ``(TOM SMITH/NEW YORK TIMES)''. We also discard captions shorter than 30 characters, which were often a series of keywords rather than an entire sentence.

\section{List of News Sources from AllSides}
\label{sec:allsides-sources}

The following are the 25 news sources in AllSides: BBC News,
Breitbart News,
CBN,
Christian Science Monitor,
CNN,
Fox News,
HuffPost,
National Review,
New York Times,
Newsmax,
NPR,
Politico,
Reason,
Reuters,
Salon,
The Guardian,
The Hill,
TheBlaze.com,
Townhall,
USA TODAY,
Vox,
Wall Street Journal,
Washington Examiner,
Washington Post,
Washington Times.

\section{Preprocessing Images for DeepFace}
\label{sec:deepface-construction}

We first perform entity linking on the text portion of AllSides articles using the Google Cloud Natural Language API,\footnote{https://cloud.google.com/natural-language/docs/analyzing-entities} identifying 9,556 entities (73K total occurrences) that have Wikipedia pages. We keep 817 entities with more than 10 occurrences in the AllSides training set and also discard entities such as \textit{God}, \textit{Jesus}, or \textit{Russians} that are clearly not politician names, resulting in a list of 722 entities. For each of these entities, we query Google Images with their name and download 50 images filtered to contain only a single face (as detected by DeepFace). Finally, we compute the similarity between each of the 50 images to select 10 reference images with the highest average similarity.

\section{Annotator Agreement with DeepFace}
\label{sec:deepface-agreement}

\begin{figure*}
    \centering
    \includegraphics[width=\textwidth]{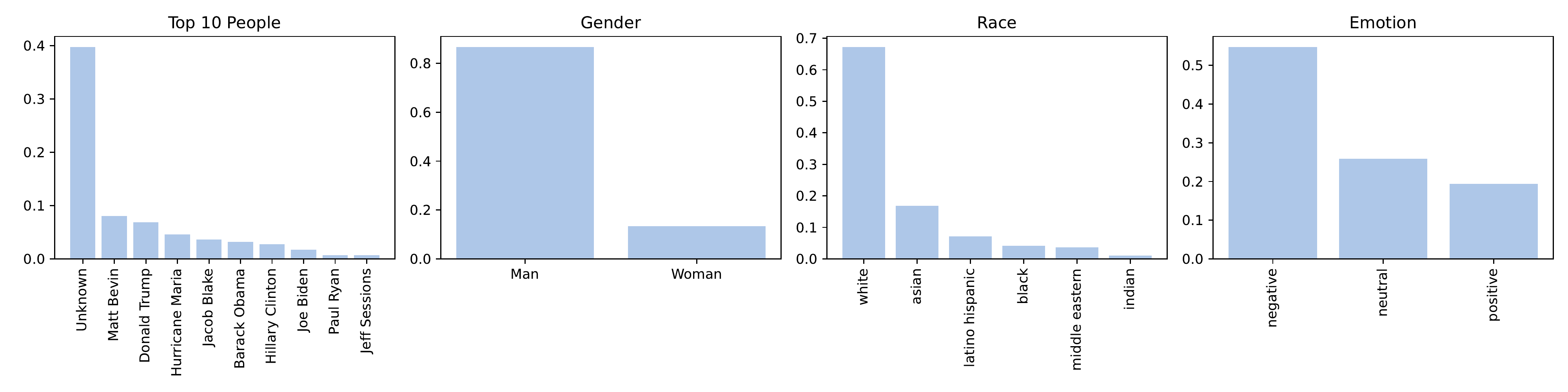}
    \caption{Other statistics of images in the AllSides dataset.}
    \label{fig:allsides-stats}
\end{figure*}

\paragraph{Agreement with DeepFace}
Because inter-annotator agreement on the pilot study was high, one author of this paper annotated images from 200 news stories following the above guidelines. We compute Cohen's kappa to measure the agreement between our annotator and the DeepFace predictions. A summary of statistics is shown in \Cref{fig:allsides-stats}. We examine each aspect of the images in turn.

For the number of people in an image, $\kappa$ = 0.69 indicates relatively high agreement. Note that we bin 6+ people into the ``group'' label. In most cases, disagreement stemmed from DeepFace recognizing more faces than the annotator; these faces were often small, blurred, or partially obscured.

\begin{figure*}
    \centering
    \includegraphics[width=\textwidth]{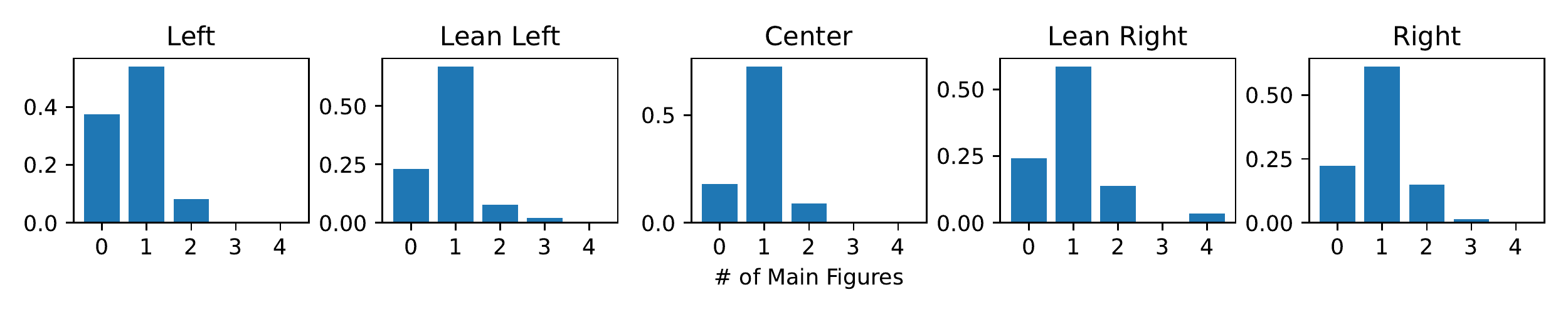}
    \caption{Number of main figures in images from different ideologies. Most images have a single main figure.
    }
    \label{fig:main-figures}
\end{figure*}

For the main figures in an image, DeepFace does not explicitly have such a notion, and we did not tell the annotator what a ``main figure'' should be. By examining the annotations, we find that main figures tend to be large, often in the center of the image, and in focus (i.e., not blurred). The number of main figures per ideology is presented in \Cref{fig:main-figures}.

DeepFace however does rank the extracted faces by their saliency. For each image, we examine the most salient face identified by the model and compare it with our annotator's labels. First, we find that identifying the figure's name results in low-annotator agreement ($\kappa$ = 0.32). We observe that DeepFace is often not able to recognize faces if they are relatively small in the image, or if the face is in side profile, even if the image is of a well-known figure. This indicates that although facial recognition tools are adept at detecting the presence of a face, they may be sensitive to the size of the face when identifying the face. We also find that DeepFace incorrectly predicts Matt Bevin (former governor of Kentucky) and Jacob Blake (a black man who was a victim of a police shooting) for many images. Due to their unusually high frequency in model predictions, we believe that these two images may represent the model's notion of a ``stereotypical'' white man and black man, respectively.

For gender, DeepFace shows high agreement ($\kappa$ = 0.76) with our annotator. Gender can be easily retrieved from a database once the person is correctly identified. Nevertheless, we find it interesting to note that DeepFace's mistakes were all misclassifying women as men; these instances included Hillary Clinton, Samantha Power, Gina Haspel, and Patty Murray.

Similar to gender, race can also be queried if the person is known. However, race is also problematic in several regards, one of which is mixed-race figures (e.g., Barack Obama). DeepFace predicts a single most apparent race, and we asked our annotator to do the same. Annotator agreement on gender prediction was relatively low ($\kappa$ = 0.34). Some common errors by DeepFace include Barack Obama and Eric Holder being misclassified as Hispanic/Latino (possibly due to skin color) and Donald Trump being misclassified as Asian (possibly due to eye size).

Emotion is difficult to annotate. Our annotator labeled fine-grained emotion for about 20\% of images, remarking that it was difficult to distinguish negative emotions. By grouping emotions into three large bins (positive, negative, neutral), DeepFace achieves moderate agreement ($\kappa$ = 0.45) with the annotator. The main disagreements were between negative and neutral emotions.

\begin{table}
    \centering
    \tiny
    \begin{tabular}{lccccc}
        \toprule
         & Left & Lean Left & Center & Lean Right & Right \\
        \midrule
        Contains Trump & 15.1\% & 9.1\% & 10.5\% & 16.2\% & 12.1\% \\
        Contains Obama & 3.5\% & 3.4\% & 1.8\% & 4.6\% & 5.5\% \\
        Contains Clinton & 3.1\% & 3.3\% & 1.2\% & 4.1\% & 3.4\% \\
        Contains Biden & 1.2\% & 2.3\% & 4.7\% & 2.9\% & 2.2\% \\
        \midrule
        With Trump & 2.4 & 1.6 & 2.3 & 1.0 & 0.7 \\
        With Obama & \textit{0.7} & 1.5 & \textit{0.0} & 1.8 & 0.6 \\
        With Clinton & \textit{1.8} & 4.0 & \textit{0.0} & 0.6 & 1.4 \\
        With Biden & \textit{0.0} & 0.9 & \textit{0.5} & 0.2 & 1.0 \\
        \bottomrule
    \end{tabular}
    \caption{Analysis of images in AllSides, focusing on the top 4 most frequent politicians. \textit{Contains NAME} indicates the percentage of images containing the specified politician. \textit{With NAME} indicates the mean number of people in images containing the specified politician, minus the specified politician; this is an indication of how often this person is pictured with other people or crowds. Italicized numbers indicate less than 10 images containing that figure, so results may not be robust.}
    \label{tab:image-figures-analysis}
\end{table}

\section{Information about Reddit and Twitter Datasets}
\label{sec:reddit-and-twitter}

\begin{table*}
    \centering
    \small

    \scalebox{0.74}{
    \begin{tabular}{ll}
    \toprule
    r/Liberal & \% \\
    \midrule
        youtube.com & 16 \\
        reddit.com & 10 \\
        rawstory.com & 3.1 \\
        businessinsider.com & 3.1 \\
        theguardian.com & 2.3 \\
        yahoo.com & 2.1 \\
        npr.org & 1.8 \\
        politico.com & 1.8 \\
        thedailybeast.com & 1.7 \\
        nbcnews.com & 1.6 \\
    \bottomrule
    \end{tabular}
    ~
    \begin{tabular}{ll}
    \toprule
    r/democrats & \% \\
    \midrule
        reddit.com & 41 \\
        youtube.com & 12 \\
        twitter.com & 3.4 \\
        i.imgur.com & 2.0 \\
        politico.com & 1.4 \\
        businessinsider.com & 1.4 \\
        nbcnews.com & 1.2 \\
        theguardian.com & 1.0 \\
        cnbc.com & 0.9 \\
        {\fontsize{5}{5}\selectfont liberalwisconsin.blogspot.com} & 0.9 \\
    \bottomrule
    \end{tabular}
    ~
    \begin{tabular}{ll}
    \toprule
    r/progressive & \% \\
    \midrule
        youtube.com & 14 \\
        reddit.com & 5.6 \\
        yahoo.com & 3.4 \\
        theguardian.com & 2.1 \\
        npr.org & 1.9 \\
        nbcnews.com & 1.8 \\
        politico.com & 1.8 \\
        twitter.com & 1.7 \\
        cnbc.com & 1.6 \\
        commondreams.org & 1.5 \\
    \bottomrule
    \end{tabular}
    ~
    \begin{tabular}{ll}
    \toprule
    r/Republican & \% \\
    \midrule
        reddit.com & 24 \\
        youtube.com & 7.8 \\
        nypost.com & 3.8 \\
        thepostmillennial.com & 3.3 \\
        timcast.com & 2.6 \\
        thefederalist.com & 2.0 \\
        redstate.com & 1.8 \\
        dailywire.com & 1.7 \\
        dailymail.co.uk & 1.6 \\
        washingtonexaminer.com & 1.5 \\
    \bottomrule
    \end{tabular}
    ~
    \begin{tabular}{ll}
    \toprule
    r/Conservative & \% \\
    \midrule
        reddit.com & 17 \\
        youtube.com & 6.1 \\
        nypost.com & 3.7 \\
        dailywire.com & 3.0 \\
        tampafp.com & 2.7 \\
        washingtonexaminer.com & 2.2 \\
        redstate.com & 2.1 \\
        dailycaller.com & 1.7 \\
        townhall.com & 1.4 \\
        thinkcivics.com & 1.4 \\
    \bottomrule
    \end{tabular}
    }

    \caption{Top 10 image sources per subreddit.}
    \label{tab:reddit-url-sources}
\end{table*}

The top 10 tweeters from each ideology, along with their tweeting stats, are shown in \Cref{tab:twitter-top10}.

\begin{table*}
    \centering
    \small
    \begin{tabular}{lrr}
        \multicolumn{3}{c}{Left} \\
        \toprule
        Username & Tweets & \% \\
        \midrule
        RepWilson & 344 & 1.3 \\
        repdinatitus & 272 & 1.0 \\
        TheDemocrats & 238 & 0.9 \\
        BillPascrell & 221 & 0.8 \\
        BruceBraley & 215 & 0.8 \\
        RepJimmyGomez & 210 & 0.8 \\
        RepDonaldPayne & 206 & 0.8 \\
        RepBeatty & 206 & 0.8 \\
        SenatorMenendez & 203 & 0.8 \\
        VP & 195 & 0.7 \\
        \bottomrule
    \end{tabular}
    ~
    \begin{tabular}{lrr}
        \multicolumn{3}{c}{Center} \\
        \toprule
        Username & Tweets & \% \\
        \midrule
        RosLehtinen & 412 & 5.9 \\
        LtGovHochulNY & 214 & 3.1 \\
        lisamurkowski & 193 & 2.8 \\
        SenatorHeitkamp & 167 & 2.4 \\
        EliseStefanik & 164 & 2.4 \\
        RepJoshG & 145 & 2.1 \\
        RepScottPeters & 141 & 2.0 \\
        SpanbergerVA07 & 140 & 2.0 \\
        RepCheri & 139 & 2.0 \\
        SenatorShaheen & 137 & 2.0 \\
        \bottomrule
    \end{tabular}
    ~
    \begin{tabular}{lrr}
        \multicolumn{3}{c}{Right} \\
        \toprule
        Username & Tweets & \% \\
        \midrule
        GovMikeDeWine & 292 & 1.2 \\
        timburchett & 268 & 1.1 \\
        GOP & 262 & 1.1 \\
        RepPeteOlson & 240 & 1.0 \\
        auctnr1 & 237 & 1.0 \\
        rep\_stevewomack & 222 & 0.9 \\
        AsaHutchinson & 206 & 0.9 \\
        LASDBrink & 204 & 0.9 \\
        NMalliotakis & 202 & 0.8 \\
        SteveWorks4You & 171 & 0.7 \\
        \bottomrule
    \end{tabular}
    \caption{Top 10 Tweeters per ideology. The percentage is percentage within their own ideology.}
    \label{tab:twitter-top10}
\end{table*}

Similar to Reddit, Twitter imposes a character limit on the length of posts (280 characters).

\section{Models}
\label{app:image-preprocessing}

\paragraph{Image Preprocessing}
Our preliminary analysis of images in our dataset shows that the majority of images contain faces of political figures. Thus, we experiment with preprocessing the image before passing it to the image encoder, based on several face-related criteria:

\begin{itemize}
\item Full Image: We leave the original image unchanged when fed to the models. If there is no face, we feed in an entirely black image (with pixel values of 0).

\item Image with Face: We leave the original image unchanged if DeepFace detects faces in the image. If DeepFace detects no faces in the image, we feed in an entirely whole black image (with pixel values of 0).

\item Only Face: If an image contains a face, we use the cropped most-salient face as indicated by DeepFace. If DeepFace detects no faces in the image, we feed in an entirely black image (with pixel values of 0).
\end{itemize}

\section{Experiments on Social Media Data}
\label{app:experiments}

Here we describe the results of our multimodal ideology prediction models on the social media data extracted from Reddit and Twitter.

\Cref{tab:reddit-exp} presents results on the Reddit dataset, in which the task is to predict either Left or Right ideology.
Overall, our VirTex-style late-fusion model with the triplet margin loss performs best.
However, accuracy is worse than on the AllSides dataset due to data mismatch: the models are trained on entire news articles, but Reddit titles are short.

\Cref{tab:twitter-exp-3} presents results on the Twitter dataset for the 3-way classification task (Left, Right, Center). We find that performance on the Center ideology is low, due to the relative lack of training data compared to Left and Right ideologies. In addition, the late-fusion multimodal models surprisingly perform worse than the text-only models. Because performance on the Center class was poor, we also experimented with a 2-way classification of Left vs. Right, whose results are shown in \Cref{tab:twitter-exp}. In this experiment, we again find that pretraining with our triplet margin loss improves performance, especially on Right-leaning tweets, over vision-only baselines.

\begin{table*}
    \centering
    \small
    \begin{tabular}{llccc}
        \toprule
                    \multicolumn{5}{c}{\textbf{Reddit}}  \\
        \textbf{Category}      & \textbf{Model}   & Overall Acc. &   Left & Right \\
        \midrule
        \multirow{2}{*}{Text-only}   & RoBERTa        &76.82 ± 0.32 &78.08 ± 2.40 & 75.62 ± 2.02 \\
        & POLITICS     &76.68 ± 1.57  &87.43 ± 3.98 &   66.36 ± 6.15\\
        Vision-only & Swin-S          &  70.90 ± 0.55 &75.05 ± 0.96  & 66.91 ± 1.82 \\
        \hdashline
        \multirow{8}{*}{Late Fusion}
        & RoBERTa+Swin-S+Cross-modal Attn.      &   & 	 \\
        &        \quad No Further Pre-training &77.79 ± 0.17  &84.88 ± 1.09 & 70.98 ± 0.72 \\
        & \quad VirTex-style+Triplet Margin  &  80.82 ± 1.05 &87.32 ± 1.09  & 74.59 ± 1.19  \\
         & POLITICS+Swin-S+Cross-modal Attn.     & \\
          &        \quad No Further Pre-training &79.06 ± 0.14 &84.65 ± 0.54  & 73.69 ± 0.24 \\
        &        \quad VirTex-style+Triplet Margin  &  \textbf{81.72 ± 0.87} &\textbf{87.79 ± 0.69} &\textbf{75.90 ± 0.70}  \\
        \bottomrule
    \end{tabular}
    \caption{Results on social media posts with both text and image from Reddit. We present mean and standard deviation over five runs. Overall, accuracy is worse than on AllSides due to data mismatch, but pretraining produces more performance gain.}
    \label{tab:reddit-exp}
\end{table*}

\begin{table*}
    \centering
    \small
    \begin{tabular}{llcccc}
        \toprule
                    \multicolumn{5}{c}{\textbf{Twitter (3-way)}}  \\
        \textbf{Category}      & \textbf{Model}   & Overall Acc. &   Left  &  Center & Right \\
        \midrule
        \multirow{2}{*}{Text-only}   & RoBERTa        &68.67 ± 0.32 &70.29 ± 2.33 & 28.58 ± 1.95 & 78.10 ± 1.51 \\
        & POLITICS     &70.32 ± 0.18  &75.29 ± 2.64 &   32.51 ± 4.16 & 76.65 ± 2.28 \\
        Vision-only & Swin-S          &  51.86 ± 0.61 &54.75 ± 4.26 & 17.47 ± 2.80  & 58.57 ± 3.57 \\
        \hdashline
        \multirow{4}{*}{Late Fusion}
        & RoBERTa+Swin-S+Cross-modal Attn.      &   & 	 \\
        &        \quad No Further Pre-training &60.95 ± 1.87 & 61.54 ± 2.91 & 22.42 ± 3.30  & 70.89 ± 3.97 \\
         & POLITICS+Swin-S+Cross-modal Attn.     & \\
          &        \quad No Further Pre-training &65.90 ± 0.63 &68.87 ± 3.83  &25.18 ± 3.32 & 74.26 ± 3.08\\
        \bottomrule
    \end{tabular}
    \caption{Results on 3-way classification of social media posts with both text and image from Twitter. We present mean and standard deviation over five runs. We find that performance on the Center ideology is low, due to the relative lack of training data. In addition, the late fusion multimodal models performed worse than text-only baselines. Because of poor performance on Center, we also present experiments on 2-way classification in \Cref{tab:twitter-exp}.}
    \label{tab:twitter-exp-3}
\end{table*}

\begin{table*}
    \centering
    \small
    \begin{tabular}{llccc}
        \toprule
                    \multicolumn{5}{c}{\textbf{Twitter (2-way, center filtered out)}}  \\
        \textbf{Category}      & \textbf{Model}   & Overall Acc. &   Left  & Right \\
        \midrule
        \multirow{2}{*}{Text-only}   & RoBERTa        &77.70 ± 0.43 &72.48 ± 1.04 & 82.45 ± 0.75 \\
        & POLITICS     &78.80 ± 0.15  &77.88 ± 1.60 &   79.67 ± 1.62 \\
        Vision-only & Swin-S          &  62.53 ± 0.85 &65.04 ± 2.72  & 60.13 ± 3.55 \\
        \hdashline
        \multirow{8}{*}{Late Fusion}
        & RoBERTa+Swin-S+Cross-modal Attn.      &   & 	 \\
        &        \quad No Further Pre-training &78.50 ± 0.06 &76.01 ± 1.61 & 80.89 ± 1.58 \\
        & \quad VirTex-style+Triplet Margin  &  79.49 ± 0.44 &\textbf{79.33 ± 2.81}  & 79.64 ± 2.57 \\
         & POLITICS+Swin-S+Cross-modal Attn.     & \\
          &        \quad No Further Pre-training &78.82 ± 0.69 &78.33 ± 1.11  & 79.29 ± 2.04\\
        &        \quad VirTex-style+Triplet Margin  &  \textbf{79.85 ± 0.18} &78.07 ± 1.70 &\textbf{81.55 ± 1.31}   \\
        \bottomrule
    \end{tabular}
    \caption{Results on social media posts with both text and image from Twitter. Here we filtered out the ``center'' ideology class as in Table \ref{tab:twitter-post-stats} and fine-tuned a 2-way classification. We present mean and standard deviation over five runs. Overall, accuracy is worse than on AllSides due to data mismatch, but pretraining gives more gains, especially on right-leaning.}
    \label{tab:twitter-exp}
\end{table*}